\documentclass{article}
\usepackage[preprint]{neurips_2025}
\usepackage[T1]{fontenc}
\usepackage[utf8]{inputenc}
\usepackage{graphicx}
\usepackage{booktabs}
\usepackage{amsmath}
\usepackage{amssymb}
\usepackage{multirow}
\usepackage{xcolor}
\usepackage{float}
\usepackage{caption}
\usepackage{tabularray}
\usepackage{placeins}
\usepackage[most]{tcolorbox}
\tcbuselibrary{listings,breakable}
\usepackage[colorlinks=true, linkcolor=blue, citecolor=blue, urlcolor=blue]{hyperref}

\newcommand{\code}[1]{{\small\texttt{#1}}}
\definecolor{goodgreen}{RGB}{0,180,0}
\definecolor{badred}{RGB}{220,30,30}

\newif\ifdraftmode\draftmodefalse
\ifdraftmode
  \providecommand{\psj}[1]{{\protect\color{red}{[Psj: #1]}}}
  \providecommand{\nishi}[1]{{\protect\color{blue}{[Nishi: #1]}}}
  \providecommand{\todo}[1]{{\protect\color{orange}{[TODO: #1]}}}
\else
  \providecommand{\psj}[1]{}
  \providecommand{\nishi}[1]{}
  \providecommand{\todo}[1]{}
\fi

\definecolor{promptbg}{RGB}{245,245,250}
\definecolor{promptframe}{RGB}{180,180,200}
\definecolor{prompttitle}{RGB}{60,60,90}

\newtcolorbox{promptbox}[1][]{
  colback=promptbg,
  colframe=promptframe,
  coltitle=white,
  colbacktitle=prompttitle,
  fonttitle=\bfseries\small,
  fontupper=\ttfamily\scriptsize,
  boxrule=0.5pt,
  arc=2pt,
  left=4pt, right=4pt, top=2pt, bottom=2pt,
  title=#1
}

\newtcblisting{promptlisting}[1]{
  breakable,
  listing only,
  colback=promptbg,
  colframe=promptframe,
  coltitle=white,
  colbacktitle=prompttitle,
  fonttitle=\bfseries\small,
  boxrule=0.5pt,
  arc=2pt,
  left=4pt, right=4pt, top=2pt, bottom=2pt,
  title=#1,
  listing options={
    basicstyle=\ttfamily\tiny,
    breaklines=true,
    breakatwhitespace=false,
    columns=fullflexible,
    keepspaces=true,
    showstringspaces=false,
  }
}

\title{Learning to Detect UI Principle Violations via Reinforcement Learning}

\author{Nishi Mehta$^1$ \quad Swathi Alse$^1$ \quad Himani Kumawat$^1$ \quad Yue Yu$^1$ \quad Pratik Jayarao$^{2}$\thanks{This work does not relate to the authors' position at Amazon.} \\
$^1$University of California, Santa Cruz \quad $^2$Carnegie Mellon University}
\date{}

\begin{document}
\maketitle

\begin{abstract}
Small language models and coding agents increasingly generate web front-end code, yet their outputs are typically evaluated primarily for functional correctness. A generated interface may compile, render, and pass unit tests while still violating established interface quality principles, including accessibility barriers, deceptive design patterns, poor visual hierarchy, and excessive decision complexity. Existing auditing approaches face a trade-off between cost, coverage, and scalability: expert human review provides rich judgment but is slow and expensive; frontier vision-language models offer broader reasoning capabilities but remain costly to deploy at scale; and rule-based tools such as axe-core and Lighthouse are inexpensive but primarily capture mechanically checkable accessibility issues.

We investigate whether a lightweight vision-language model can serve as an effective critic for generated interfaces. We unify 19 interface-quality principles from three complementary sources of HCI knowledge: WCAG 2.2 accessibility standards, deceptive design taxonomies, and established theories of perception, cognition, and interaction. To train this critic, we construct a verified dataset of approximately 10,000 generated web pages by synthetically injecting known violations into clean, LLM-generated Tailwind pages. A frontier teacher model verifies that each injected violation is visually observable before the sample is retained.

Continued reinforcement learning on a 4B vision-language model improves micro-F1 from 36\% to 84\%, with 13 of 19 principles exceeding 80\% F1. The resulting critic can audit generated interfaces, filter low-quality interface training data, and provide a reward signal for design-aware code generation. We release our data-generation recipe and injection/verification prompts to support reproducible evaluation and future work on scalable interface-quality assessment.
\end{abstract}

\section{Introduction}
\label{sec:intro}

Language models and coding agents now generate a large and growing share of web front-end code. These models are trained largely on internet-scale code and optimized for functional correctness, often with reinforcement learning against unit tests and execution feedback, but they receive very little explicit supervision on interface quality. With little of this feedback in the loop, a generated page can compile, render, and pass every unit test while still embedding poor design choices: functional correctness says nothing about \emph{design} quality. As a result, generated pages may violate well-documented UI/UX principles, spanning three categories of failure: accessibility barriers that exclude users with disabilities \citep{wcag22}, dark patterns that manipulate users \citep{mathur2019darkpatterns}, and violations of cognitive usability principles that degrade usability for everyone \citep{nielsen1994}.

Two characteristics of how these models are built make such failures likely. \textbf{Training data.} They are trained on internet-scale web data in which manipulative patterns are common, so a model can reproduce a dark pattern as an ordinary design convention rather than treat it as something to avoid. \textbf{Limited visual feedback.} Many code-generating models operate on text alone and never render the page they produce, so violations that surface only visually, such as low contrast, poor visual hierarchy, inconsistent spacing, or misaligned layouts, can pass unnoticed at generation time. As automated code generation scales, automatically auditing generated pages for these violations becomes increasingly important.

Common auditing options each trade off cost against coverage. Expert human review is thorough but slow and expensive. Frontier LLMs are capable but costly to run at the scale generated code demands. Rule-based checkers (axe-core, Lighthouse, WAVE) are cheap but tend to focus on mechanical accessibility rules, and are less suited to dark patterns or cognitive violations \citep{aidui2023, scrapeai2024}. We explore a trained small vision-language model as a UI/UX \emph{critic}: a middle ground that is cheap enough to run on generated pages, yet accurate after lightweight training.

Training such a model requires labeled data, which is costly to obtain at scale: annotating principle violations across diverse pages demands expert judgment, and the space of principles, page types, and severities is large. We address this through controlled injection. Rather than label existing pages, we introduce known violations into clean pages, yielding exact per-principle labels by construction; a teacher model then verifies that each injected violation is genuinely present before the sample is retained. Because each page is paired with diverse, controlled violations, the model is trained to recognize the underlying principles rather than memorize injection templates. Beyond standalone auditing, we propose such a critic as a reusable component: it can gate or flag violations in automated UI testing, filter UI/UX training data, or provide a reward signal for design-aware code generation.

\begin{figure*}[t]
\centering
\includegraphics[width=\textwidth]{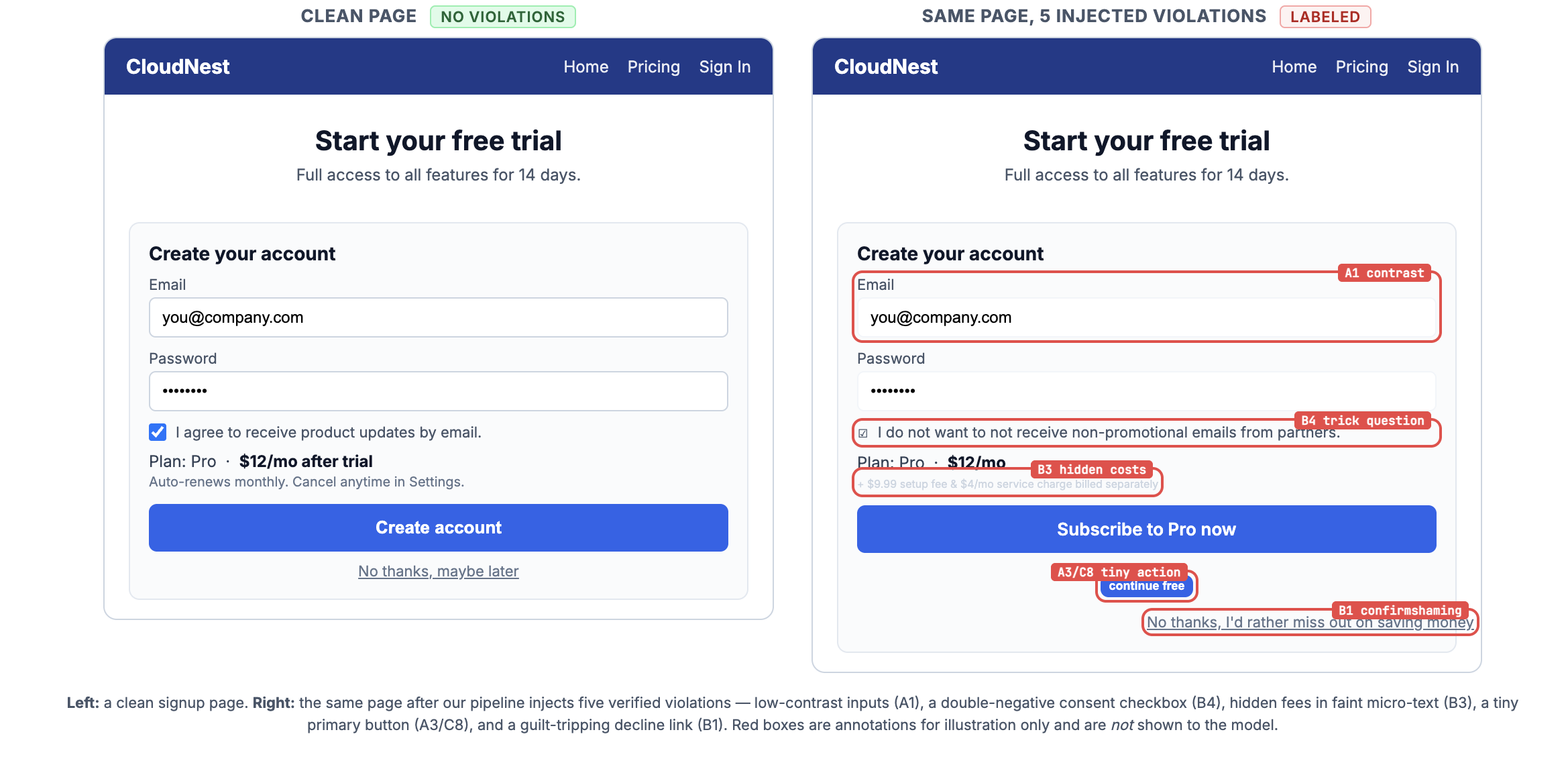}
\caption{Our task. \textbf{Left:} a clean signup page. \textbf{Right:} the same page after our pipeline injects five \emph{verified} violations, low-contrast inputs (A1), a double-negative consent checkbox (B4), hidden fees in faint micro-text (B3), a tiny primary action (A3/C6), and a guilt-tripping decline link (B1). Red boxes are annotations for illustration only and are not shown to the model, which receives the rendered screenshots and the HTML source and must output a binary label per principle.}
\label{fig:example}
\end{figure*}

Our contributions:
\begin{enumerate}
    \item \textbf{Unified violation taxonomy.} We consolidate 19 UI principles spanning three established HCI frameworks (WCAG 2.2 accessibility, dark pattern ontologies, and cognitive and perceptual design laws) into a single detection taxonomy grounded in citable prior work.
    \item \textbf{Verified-injection dataset.} We build a multi-pass pipeline that generates clean Tailwind pages and injects diverse violations through a staged \emph{inject/re-render/verify} loop, in which a teacher model confirms every injected violation is genuinely present and perceptible before it is kept, yielding a balanced, verified set of roughly 10K labeled pages. We release the full recipe and the injection/verification prompts so the dataset can be reproduced.
    \item \textbf{A small, trainable detector.} A single 4B vision-language model, after lightweight training, detects violations across accessibility, dark patterns, and cognitive and perceptual principles at a fraction of a frontier LLM's inference cost, making per-page auditing of generated interfaces practical.
\end{enumerate}

\section{Related Work}
\label{sec:related}
\paragraph{Dark Pattern Detection.} The deceptive design detection literature is predominantly supervised. AIDUI \citep{aidui2023} uses object detection on mobile app screenshots to classify dark patterns. ScrapeAI \citep{scrapeai2024} combines HTML DOM parsing with LLMs in a multi-modal framework. Several works apply BERT-family models to textual dark pattern classification in UI copy. All are limited to dark patterns alone; none addresses accessibility or cognitive principles.

\paragraph{Accessibility Testing.} Automated accessibility testing is dominated by rule-based tools (axe-core, Lighthouse, WAVE) that check a subset of WCAG success criteria programmatically. These achieve high precision on the rules they cover but cannot detect contextual violations (e.g., error identification requires understanding what constitutes an ``error explanation''). A11YN \citep{a11yn2025} uses RL with WCAG-based rewards but for \textit{generating} accessible code, not detecting violations. We are not aware of prior work that trains a model to detect accessibility violations via RL.

\paragraph{RL for UI.} WebGen-R1 \citep{webgenr1_2026} and GUI-Genesis \citep{guigenesis2026} apply RL to website generation and GUI-agent training respectively, not violation detection. The closest paradigm is RLVR for mathematical reasoning \citep{deepseekmath2024, deepseekr1}, where verifiable rewards enable RL without human labels; we adapt it to UI principles. Our setting differs from all of these in combining RL training, raw HTML/CSS input, violation detection, coverage across all four principle families, and verifiable rewards from synthetic injection.

\section{Violation Taxonomy}
\label{sec:taxonomy}

\begin{figure*}[t]
\centering
\includegraphics[width=\textwidth]{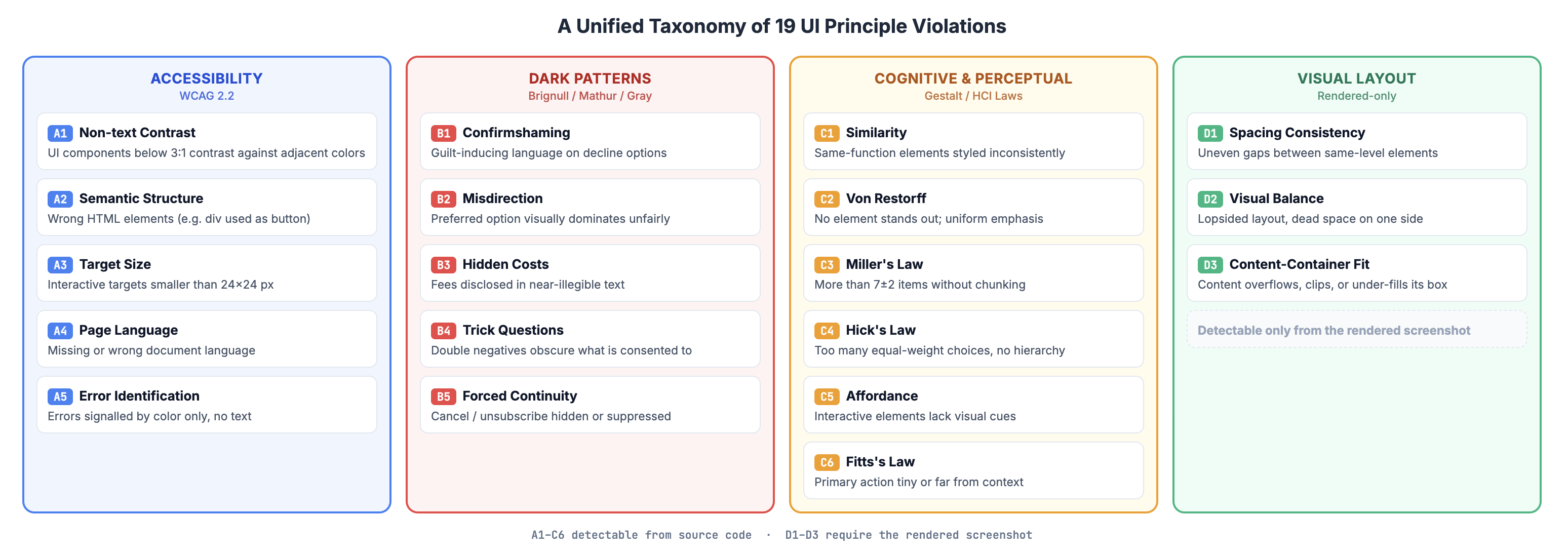}
\caption{The 19-principle violation taxonomy, grouped into four families. Principles A1--C6 are detectable from the HTML source; D1--D3 require the rendered screenshot.}
\label{fig:taxonomy}
\end{figure*}

We construct a unified taxonomy of 19 UI principle violations drawn from three established frameworks in the HCI and accessibility literature (Figure~\ref{fig:taxonomy}): five WCAG 2.2 \textbf{accessibility} criteria (A1--A5), five \textbf{dark patterns} (B1--B5), six \textbf{cognitive and perceptual} principles (C1--C6), and three \textbf{visual composition} principles (D1--D3). Each principle is selected to satisfy three properties: (1)~\textit{groundedness}, it traces to a citable standard or empirical finding; (2)~\textit{injectability}, the violation can be introduced into a static HTML/CSS document through programmatic transformation or targeted LLM editing; and (3)~\textit{verifiability}, its presence can be confirmed from the rendered page or its source. We give each principle a precise definition and a concrete violation example below; Table~\ref{tab:defs} (Appendix~\ref{app:defs}) collects them.

\subsection{Principle Definitions}

\paragraph{Accessibility (A1--A5).} These derive from WCAG 2.2 success criteria \citep{wcag22}, the de facto standard for web accessibility. \textbf{A1 Non-text Contrast} (SC 1.4.11) requires that UI components such as borders, icons, and control outlines maintain at least a 3:1 contrast ratio against adjacent colors; a violation is, for example, input borders faded to near-white so a low-vision user cannot find the field boundary. \textbf{A2 Semantic Structure} requires that the correct semantic element be used for its role, since assistive technology relies on it; a violation is a clickable \code{<div>} styled to look like a button, or an input with no associated \code{<label>}. \textbf{A3 Target Size} (SC 2.5.8) requires interactive targets to be at least $24\times24$ px; a violation is a primary action shrunk to a few pixels tall. \textbf{A4 Page Language} (SC 3.1.1) requires the document to declare its language; a violation is a missing or incorrect \code{lang} attribute, which breaks screen-reader pronunciation. \textbf{A5 Error Identification} (SC 3.3.1) requires form errors to be described in text rather than signaled by color alone; a violation is an invalid field marked only by a red border with no message.

\paragraph{Dark patterns (B1--B5).} These draw on the canonical deceptive-design references \citep{brignull2010, mathur2019darkpatterns, gray2024ontology}, and their defining feature is intent: the interface is shaped to benefit the operator at the user's expense. \textbf{B1 Confirmshaming} uses guilt-inducing language on a decline option, e.g.\ a dismiss link reading ``No thanks, I'd rather pay full price.'' \textbf{B2 Misdirection} uses visual emphasis to steer the user toward the option that benefits the business, e.g.\ a prominent ``Subscribe'' button beside a near-invisible ``No thanks'' link. \textbf{B3 Hidden Costs} discloses fees late or in near-illegible text, e.g.\ a ``+\$9.99 setup fee'' set in tiny faint type next to the total. \textbf{B4 Trick Questions} phrases consent confusingly, e.g.\ a double negative (``Uncheck this box if you do not want to not receive emails''). \textbf{B5 Forced Continuity} hides or suppresses the path to cancel or unsubscribe, e.g.\ a subscription page with no visible cancellation control.

\paragraph{Cognitive and perceptual (C1--C6).} These operationalize Gestalt grouping principles \citep{nielsen1994} and quantitative HCI laws \citep{miller1956, hick1952, fitts1954}, and their violations are relative to the page's own visual language rather than absolute. \textbf{C1 Similarity} holds that elements with the same function should look alike; a violation styles equivalent navigation links inconsistently so they appear unrelated. \textbf{C2 Von Restorff} (the isolation effect) holds that the most important item should stand out; a violation gives every call-to-action identical emphasis so nothing draws the eye. \textbf{C3 Miller's Law} draws on Miller's classic observation regarding working-memory capacity: interfaces should organize information into manageable chunks rather than long undifferentiated lists; a violation presents a flat 15-item menu with no grouping. \textbf{C4 Hick's Law} holds that interfaces should reduce unnecessary decision complexity, since decision time grows with the number of equal-weight choices, so a hierarchy should be imposed; a violation presents a dozen equally-weighted buttons with no primary action. \textbf{C5 Affordance} holds that interactive elements should provide clear signifiers of interactivity; a violation is a clickable card with no hover, cursor, or other cue. \textbf{C6 Fitts's Law} holds that a target is harder to hit the smaller and farther away it is, so primary actions should be large and near their context; a violation places a tiny submit button far from the form it submits.

\paragraph{Visual Composition (D1--D3).} These capture rendering-level quality that is visible only in the screenshot, not recoverable from source alone. \textbf{D1 Spacing Consistency} requires even gaps between same-level elements; a violation is a card grid with jittery, unequal spacing. \textbf{D2 Visual Balance} requires visual weight to be distributed rather than lopsided; a violation crams all content to one side and leaves the other as dead space. \textbf{D3 Content-Container Fit} requires content to fit its container without overflow, clipping, or under-fill; a violation is heading text spilling out of its button.

\subsection{Scope and Detection Difficulty}

The families form a gradient of detection difficulty. A2--A4 are close to mechanically checkable from the markup; A1, A5, and the dark patterns require reading content and weighing intent; the cognitive and visual-layout principles require a \emph{relative} judgment over the rendered page, comparing elements to one another rather than inspecting one in isolation. This spread is deliberate: it lets a single detector be measured across the full range from rule-like to perceptual. The visual-layout family (D1--D3) in particular is detectable only from the screenshot, which is what motivates a multimodal rather than source-only model.

We deliberately exclude principles requiring runtime behavior (focus trapping, keyboard navigation), cross-page consistency, or user-study metrics such as learnability and satisfaction. The 19 principles are the set we can inject into, and reliably verify from, a single static HTML/CSS document together with its rendered screenshots.

\section{Data}
\label{sec:data}
\begin{figure*}[t]
\centering
\includegraphics[width=\textwidth]{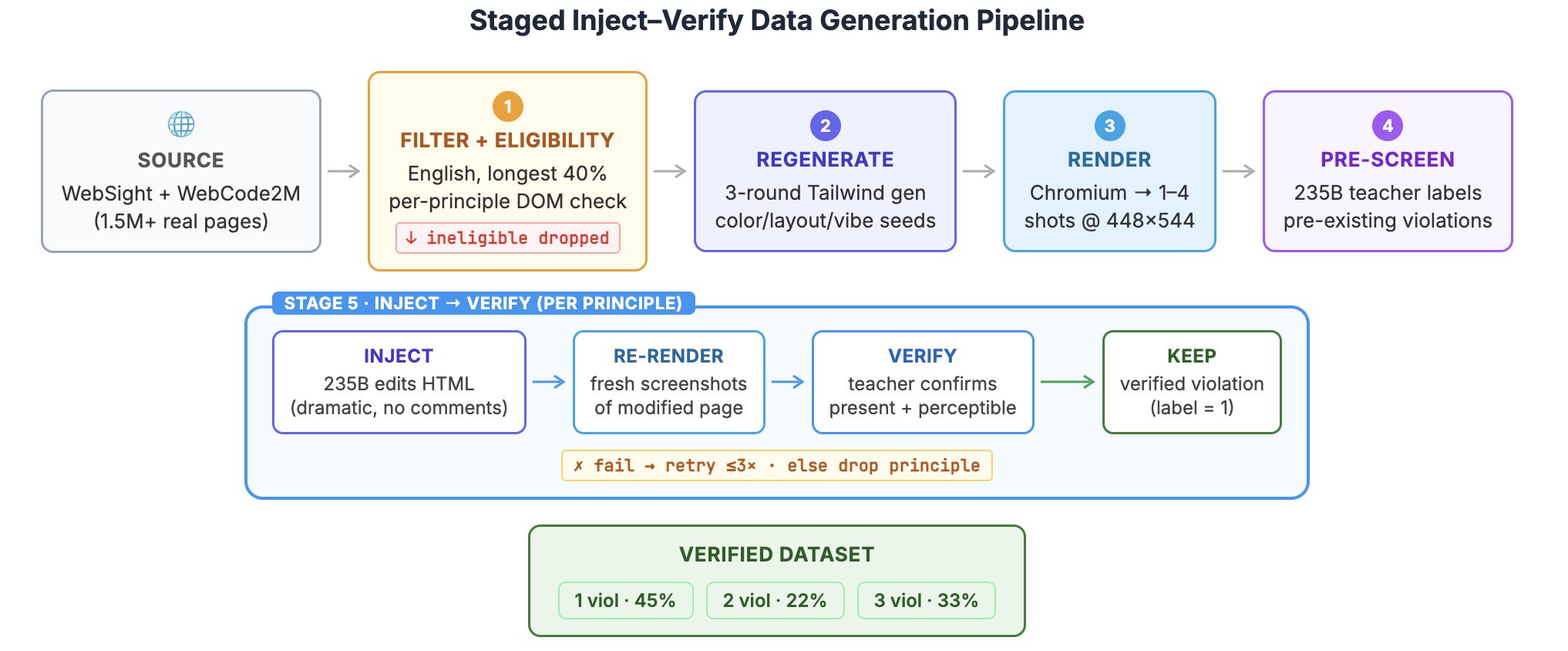}
\caption{The data generation pipeline. Real pages are filtered and screened for per-principle eligibility, regenerated as clean Tailwind documents, rendered, and pre-screened for organic violations. The core is the staged inject$\rightarrow$re-render$\rightarrow$verify loop (Stage 5): the teacher edits the HTML, we re-render, and the teacher must confirm the violation is genuinely present and perceptible before it is kept (else retry up to 3$\times$, or drop). Each page accumulates 1--3 verified violations.}
\label{fig:pipeline}
\end{figure*}

Training a reliable critic requires data whose labels we can trust. Hand-labeling principle violations across diverse pages is expensive and subjective, so we instead synthesize labeled data: we inject known violations and have a strong teacher model, Qwen3-VL-235B-A22B, which sees both the rendered screenshots and the HTML, confirm each one. Our pipeline has two parts, described in the following subsections. First, we \emph{generate} a diverse pool of clean, realistic web pages (Section~\ref{sec:source-prep}). Second, we \emph{inject} known violations into these pages and verify each one (Section~\ref{sec:injection}). Generating the violations ourselves makes the label exact by construction, since we know precisely what was injected, and the verification step keeps only violations the teacher confirms are genuinely present, so we trust the resulting labels with high confidence. Figure~\ref{fig:pipeline} shows the full pipeline.

\subsection{Page Generation}
\label{sec:source-prep}

Generating pages by prompting an LLM from scratch tends to collapse toward a few templates, yielding low visual and structural diversity. To avoid this, we seed generation from real pages. We start from \textbf{WebCode2M} \citep{webcode2m}, a large corpus of scraped web pages that are structurally complex, with forms, multi-level navigation, and diverse interactive elements, and use each real page as a structural seed for a clean regenerated page.

\paragraph{Domain and page-type coverage.} Real corpora are heavily skewed toward a few page kinds, so uniform sampling would leave categories where violations matter most (for example checkout or login pages, where dark patterns concentrate) badly underrepresented. We therefore classify every seed page along two axes with an LLM: \emph{domain} (20 categories such as e-commerce, healthcare, finance, gaming) and \emph{page type} (20 categories such as login, checkout, pricing, dashboard); the full lists are in Appendix~\ref{app:gen-categories}, Table~\ref{tab:categories}. We then stratify-sample across both axes, capping over-represented categories and retaining all of the rare ones, so the working set spans the full grid of (domain, page type) combinations rather than the corpus's natural skew.

\paragraph{Regeneration.} Each seed page is rebuilt as a clean, self-contained Tailwind document through a three-round generation process (prompts in Appendix~\ref{app:gen-prompts}): round 1 produces a domain-specific design plan, round 2 implements it as HTML conditioned on sampled color, layout, and ``vibe'' seeds for visual diversity, and round 3 enriches the page with additional domain-appropriate sections and interactions. Each round increases complexity, so the regenerated pages are richer than their seeds and provide more substrate into which a wider range of violations can be injected. We render each page with headless Chromium to 1--4 screenshots, downscaled to $448\times544$ (238 vision tokens each). Filtering details (English-only, length) are in Appendix~\ref{app:pipeline}.

\paragraph{Eligibility and pre-screening.} Not every page can host every violation, so before injection we parse each page and compute a per-principle eligibility vector: a page receives a candidate violation only if its DOM can support it (for instance, a trick-question requires a form, and misdirection requires at least two action elements); pages supporting none are discarded (full rules in Appendix~\ref{app:pipeline}). Because the regenerated pages also already contain organic violations, we then pre-screen each one with the teacher, which labels which of the 19 principles are already violated. Principles it marks clean become scorable negatives, while pre-existing violations we do not inject are excluded from scoring, so the critic is not penalized for flagging an organic issue we never labeled.

\subsection{Injecting and Verifying Violations}
\label{sec:injection}

Once a page has been generated, screened for eligibility, and pre-screened for organic violations, we inject the target violations. A natural approach would be to ask the teacher to edit the page and trust that it did so, but in practice this is unreliable: asked to introduce a violation, the model often adds an inert comment, dummy text, or a change too subtle to perceive while still reporting success. We therefore treat each injection as a propose-and-verify step and keep a violation only once the teacher confirms it is genuinely present.

Concretely, for each target principle we run three steps. \textit{Inject:} the teacher receives the current screenshots, the full HTML, and a principle-specific instruction, and returns the modified HTML together with a separate natural-language explanation, which we keep out of the code so that no telltale comments leak into the page. \textit{Re-render:} we render the modified HTML to fresh screenshots, because a violation is only useful if it is actually visible in the rendered page, not merely present in the markup. \textit{Verify:} the teacher compares the original and modified HTML and screenshots and judges whether the intended violation is now genuinely present and perceptible. If verification fails we retry the injection up to three times, and a principle that still does not verify is dropped for that page. Each principle has its own injection and verification prompt, written to make the violation unambiguous (we use deterministic transforms where a violation is fully mechanical, such as altering the \code{lang} attribute for page-language); all prompts are listed in Appendix~\ref{app:inject-prompts}.

\paragraph{Staged accumulation.} To produce pages with varying numbers of violations, injection runs in three stages: stage~1 adds one verified violation to every page, stage~2 a second to a subset, and stage~3 a third, carrying earlier violations forward (re-rendered and re-verified together) with per-page de-duplication and balanced sampling across principles. Taking each page at its deepest successful stage yields 1--3 verified violations per page ($\approx$45\%, 22\%, and 33\% for one, two, and three violations respectively).

\subsection{Ground Truth and Reward}

The pipeline gives deterministic ground truth with \textit{no judge at training time}: injected-and-verified principles are positives (label 1), pre-screen-confirmed-clean principles are negatives (label 0), and pre-existing-but-not-injected violations are excluded. The reward is the per-sample $F_1$ of the positive class over the scorable set $S$ (injected $\cup$ clean):
\begin{equation}
    R = F_1 = \frac{2\,\mathrm{TP}}{2\,\mathrm{TP} + \mathrm{FP} + \mathrm{FN}},
\end{equation}
counted over $k\in S$. We use $F_1$ over accuracy because $S$ is class-imbalanced (1--3 positives among $\sim$10--18 scored), so accuracy would reward predicting ``clean'' everywhere. An unparseable \code{<labels>} block scores $R=0$, giving a strong format-compliance signal.

\section{Training Setup}
\label{sec:training}

\subsection{Model}

We use Qwen3-VL-4B-Thinking as the student model, a 4.6B-parameter vision-language model (4B text transformer + 600M vision encoder) supporting both text and image inputs natively. The ``Thinking'' variant produces chain-of-thought reasoning in \code{<think>...</think>} blocks before outputting structured labels. For injection, verification, and pre-screening, we use Qwen3-VL-235B-A22B as the teacher, served via vLLM across 8$\times$H200 GPUs.

\subsection{Multimodal Input}

Each training sample consists of: (1) rendered screenshots of the page (1--4 images at 448$\times$544 pixels, producing 238 vision tokens per image after patch embedding with patch\_size=16 and merge\_size=2), and (2) the full HTML source code as text. The model receives both modalities: screenshots enable detection of visual-only violations (D1--D3) invisible in source code, while the HTML enables structural violations (A2, A4) not visible in the rendered output.

\subsection{Algorithm}

We use GRPO \citep{deepseekmath2024} with 16 samples per prompt and no KL penalty ($\lambda_{\text{KL}} = 0$). The reward is fully deterministic, so no LLM judge is needed during training. The multi-label $F_1$ reward is dense: a rollout that recovers some but not all injected violations, or that over-flags a clean principle, receives a partial score rather than all-or-nothing feedback, giving the policy a smooth gradient toward both higher recall and higher precision.

\subsection{Task Format}

The model receives an HTML/CSS page and outputs structured judgments:

\begin{promptbox}[Model Input]
\rmfamily
You are a strict UI/UX auditor. You are shown screenshot(s) of a rendered web page followed by its HTML source code. Identify which of the 19 UI principles the page violates. For each principle, output 1 if the page violates it, or 0 if it does not.\\[4pt]
\ttfamily
SCREENSHOT(S):\\
<image> ... <image>\\[2pt]
HTML SOURCE:\\
\{html\_content\}\\[4pt]
\rmfamily
Output your answer inside <labels> tags, one principle per line:\\
\ttfamily
<labels>\\
A1\_non\_text\_contrast: 0\\
A2\_semantic\_structure: 0\\
...\\
D3\_content\_container\_fit: 0\\
</labels>
\end{promptbox}

The thinking model first reasons in a \code{<think>} block (analyzing the page principle-by-principle), then outputs the structured \code{<labels>} block with a binary 1/0 for each of the 19 principles. RL learns to produce concise reasoning followed by correct predictions, terminating the \code{<think>} block in time to emit a parseable label set.

\section{Experiments}
\label{sec:experiments}

\subsection{Setup}

\paragraph{Data.} We train on 9,251 pages with verified injected violations (1--3 per page) and evaluate on 500 held-out pages, stratified by violation count and split so that all instances of a given source page lie on the same side, avoiding train/eval leakage. Inputs comprise the page's screenshots and HTML and fit within a 32K-token budget.

\paragraph{Training.} We perform continued RL from the post-trained Qwen3-VL-4B-Thinking checkpoint, optimizing the $F_1$ reward with GRPO (32 prompts per rollout batch, 16 samples per prompt, constant learning rate $10^{-6}$, $\epsilon$-clip $[0.2, 0.28]$, sampling temperature 1.0, response length up to 8K tokens). We train until the eval reward converges and report the converged checkpoint; the baseline is the same checkpoint evaluated zero-shot.

\paragraph{Evaluation metrics.} We report (1)~format compliance, the fraction of samples producing a valid \code{<labels>} block; (2)~micro-averaged precision, recall, and $F_1$ on the positive (violation-present) class; and (3)~per-principle $F_1$, all over the 500 held-out eval pages. Samples with no parseable output count as predicting 0 for every principle. We report $F_1$ as a percentage throughout.

\section{Results}
\label{sec:results}

We report micro-averaged precision, recall, and $F_1$ over the 500-page eval set (Table~\ref{tab:main}), and per-principle $F_1$ (Table~\ref{tab:perprinciple}), for the model before training (zero-shot) and after continued RL. All numbers are percentages.

\begin{table}[t]
\centering
\small
\begin{tblr}{
  colspec = {l cccc},
  hline{1} = {1.5pt, solid},
  hline{Z} = {1.5pt, solid},
  hline{2} = {0.8pt, solid},
  row{1} = {font=\bfseries},
  row{3} = {font=\bfseries},
  column{1} = {bg=gray!10},
}
  Stage & Precision & Recall & F1 & Format \\
  Zero-shot & 44\% & 30\% & 36\% & 80\% \\
  +RL & 94\% & 76\% & 84\% & 96\% \\
\end{tblr}
\caption{Micro-averaged detection performance (\%) on the 500-page eval set. Format is the fraction of responses emitting a valid \code{<labels>} block. The trained (+RL) row is shown in bold.}
\label{tab:main}
\end{table}

\subsection{Zero-Shot Performance}

Before training, the model is uneven rather than uniformly weak (Table~\ref{tab:perprinciple}, Base). It handles violations whose evidence is explicit in the text it reads: hidden costs (B3, 77\%) are spelled out in the markup, and trick questions (B4) and confirmshaming (B1), which hinge on the wording of a label or button, are partly caught because the model reads that wording directly. Semantic-structure issues (A2), visible in the tag names themselves, are likewise above half.

The zero-shot failures follow two patterns, and several are the violations a deployed auditor would most want flagged. First, principles that require \emph{looking} at the rendered page rather than reading the source are near zero: non-text contrast (A1) and content-container fit (D3) both sit at $0\%$, and uneven spacing (D1, 4\%) is barely detected, even though the page is supplied as an image, which indicates the untrained model leans on the HTML and largely ignores the screenshot. Second, principles that require a \emph{relative} judgment across elements fail: misdirection (B2, 3\%), which depends on one option visually dominating another, and Fitts's Law (C6, 3\%), which depends on a primary action being too small or far from its context, are almost entirely missed. The manipulative dark patterns are the most consequential of these: a zero-shot critic that misses misdirection would pass the pages it is most important to catch.

\subsection{Main Results}

Continued RL raises micro-$F_1$ from 36\% to 84\% and recall from 30\% to 76\% (Table~\ref{tab:main}), and the per-principle view (Table~\ref{tab:perprinciple}, +RL) shows the gain is broad rather than concentrated: 13 of 19 principles exceed 80\%, and every principle family improves. The visual-only principles that the zero-shot model ignored improve the most, from near zero to high accuracy: spacing (D1) from 4\% to 94\%, content-container fit (D3) from 0\% to 92\%, and non-text contrast (A1) from 0\% to 64\%. This is consistent with the model learning to use the rendered screenshot rather than the HTML alone, since these violations are not reliably recoverable from source. Dark patterns and accessibility cues with textual evidence saturate near the top, with confirmshaming (B1), hidden costs (B3), and trick questions (B4) all at or above 97\%.

Format compliance rises from 80\% to 96\% over training. The zero-shot thinking model often over-reasons, exhausting its token budget before emitting a label block; such responses are unparseable and score zero, which is the main reason zero-shot recall (30\%) trails zero-shot precision (44\%). Because the reward is zero for any unparseable response, the policy learns to reason concisely and terminate in time, which both recovers recall and lowers inference cost.

\begin{table}[t]
\centering
\small
\begin{tblr}{
  colspec = {l l cc | l l cc},
  colsep = 5pt,
  hline{1} = {1.5pt, solid},
  hline{Z} = {1.5pt, solid},
  hline{2} = {0.8pt, solid},
  row{1} = {font=\bfseries},
  column{4} = {font=\bfseries},
  column{8} = {font=\bfseries},
}
  Cat & Principle & Base & +RL & Cat & Principle & Base & +RL \\
  A1 & Contrast            & 0\% & 64\% & C1 & Similarity        & 43\% & 90\% \\
  A2 & Semantic Structure  & 58\% & 92\% & C2 & Von Restorff      & 16\% & 75\% \\
  A3 & Target Size         & 42\% & 84\% & C3 & Miller's Law      & 49\% & 54\% \\
  A4 & Page Language       & 44\% & 86\% & C4 & Hick's Law        & 33\% & 76\% \\
  A5 & Error ID            & 31\% & 89\% & C5 & Affordance        & 49\% & 94\% \\
  B1 & Confirmshaming      & 53\% & 97\% & C6 & Fitts's Law       & 3\% & 36\% \\
  B2 & Misdirection        & 3\% & 55\% & D1 & Spacing           & 4\% & 94\% \\
  B3 & Hidden Costs        & 77\% & 97\% & D2 & Visual Balance    & 39\% & 99\% \\
  B4 & Trick Questions     & 62\% & 100\% & D3 & Container Fit     & 0\% & 92\% \\
  B5 & Forced Continuity   & 29\% & 95\% & & & & \\
\end{tblr}
\caption{Per-principle $F_1$ (\%) on the 500-page eval set. Base = zero-shot, +RL = after continued RL. RL lifts 13 of 19 principles above 80\%; the hardest remaining are misdirection, Miller's Law, and Fitts's Law.}
\label{tab:perprinciple}
\end{table}

\subsection{Error Analysis}

Three principles remain hard after training: Fitts's Law (C6, 36\%), Miller's Law (C3, 54\%), and misdirection (B2, 55\%). They share a structure that separates them from the principles the model masters: each requires a \emph{relative} judgment across multiple elements rather than inspection of a single element in isolation. Misdirection compares the visual weight of two competing actions, Fitts's Law relates a target's size and position to its surrounding context, and Miller's Law counts grouped items against a working-memory threshold. At the $448\times544$ training resolution, these fine-grained spatial and counting comparisons are difficult to make reliably. We regard them as the frontier for a small UI/UX critic rather than a limitation of the training procedure.

\FloatBarrier
\section{Conclusion}
\label{sec:conclusion}

As language models and coding agents generate an increasing share of web interfaces, auditing those interfaces for UI/UX principle violations becomes a practical need that human review and frontier LLMs are too costly to meet at scale. We presented a small vision-language model trained to serve as this critic, covering 19 principles spanning WCAG 2.2 accessibility, dark patterns, and cognitive and perceptual design laws in a single model. To train it without expensive annotation, we generated a diverse, domain-balanced pool of pages and injected violations that a teacher model verifies are genuinely present, yielding labels we trust by construction. Continued RL raises micro-$F_1$ from \textbf{36\%} to \textbf{84\%}, with \textbf{13 of 19} principles exceeding 80\%; the gains are largest on exactly the violations a zero-shot model misses, the visual and manipulative ones, while relational judgments such as misdirection and Fitts's Law remain the frontier. Cheap enough to run on every generated page, such a critic can audit interfaces, filter UI/UX training data, or provide a design-aware reward signal for code generation. We release our data-generation recipe and prompts so the dataset can be reproduced and extended.

\section*{Limitations}

\paragraph{Single run.} Due to compute constraints we report a single model trained with a single random seed; we do not characterize variance across seeds or runs.

\paragraph{Teacher-derived labels, no human validation.} Both the training ground truth and the evaluation labels are produced by a teacher model (injection, verification, and pre-screening) rather than human annotators, so they may contain errors, and the reported gains are measured against teacher-derived labels rather than a human-verified test set. A human-labeled evaluation would give a more trustworthy estimate of real-world accuracy.

\paragraph{Resolution ceiling on relational judgments.} The three hardest principles after training, misdirection (B2), Miller's Law (C3), and Fitts's Law (C6), all require relative comparisons across elements (relative visual weight, item counts, or size-and-distance to context). At the $448\times544$ training resolution these fine-grained spatial and counting comparisons are difficult to make reliably, which likely caps performance on this family. Higher-resolution inputs or explicit region cues may be needed to close the gap.

\paragraph{Synthetic-injection scope.} The critic is trained and evaluated on injected violations in regenerated Tailwind pages. Whether it transfers to organically-occurring violations in real production interfaces, which may differ in style and co-occurrence patterns, is not directly measured here.

\bibliography{custom}
\bibliographystyle{plainnat}

\appendix

\section{Taxonomy Definitions}
\label{app:defs}

Table~\ref{tab:defs} gives a precise definition and an example violation for each of the 19 principles (Section~\ref{sec:taxonomy}).

\begin{table*}[h]
\centering
\small
\begin{tblr}{
  width = \textwidth,
  colspec = {l l X[2.1,l] X[2.4,l]},
  hline{1} = {1.5pt, solid},
  hline{Z} = {1.5pt, solid},
  hline{2} = {0.8pt, solid},
  hline{7,12,18} = {0.4pt, solid},
  row{1} = {font=\bfseries},
  column{1} = {bg=gray!10},
}
  ID & Principle & Definition & Example violation \\
  A1 & Non-text Contrast    & UI components keep $\geq$3:1 contrast against adjacent colors (WCAG 1.4.11) & Input borders faded to near-white; field boundary invisible \\
  A2 & Semantic Structure   & Correct semantic element used for its role & Clickable \code{<div>} styled as a button; input with no \code{<label>} \\
  A3 & Target Size          & Interactive targets $\geq 24\times24$ px (WCAG 2.5.8) & Primary action shrunk to a few pixels tall \\
  A4 & Page Language        & Document declares its language (WCAG 3.1.1) & Missing or wrong \code{lang} attribute on \code{<html>} \\
  A5 & Error Identification  & Errors described in text, not color alone (WCAG 3.3.1) & Invalid field shown only by a red border, no message \\
  B1 & Confirmshaming       & Decline options free of guilt-inducing language & ``No thanks, I'd rather pay full price'' \\
  B2 & Misdirection         & Visual emphasis does not steer toward the operator's interest & Bright ``Subscribe'' beside a near-invisible ``No thanks'' \\
  B3 & Hidden Costs         & Fees disclosed clearly and on time & ``+\$9.99 setup fee'' in tiny faint type by the total \\
  B4 & Trick Questions      & Consent phrased unambiguously & Double negative: ``Uncheck if you don't want to not receive\dots'' \\
  B5 & Forced Continuity    & Cancel / unsubscribe remains reachable & Subscription page with no visible cancel control \\
  C1 & Similarity           & Same-function elements look alike & Equivalent nav links styled inconsistently \\
  C2 & Von Restorff         & The important item stands out & Every call-to-action given identical emphasis \\
  C3 & Miller's Law         & Information organized into manageable chunks, not long undifferentiated lists & Flat 15-item menu with no grouping \\
  C4 & Hick's Law           & Equal-weight choices kept few; hierarchy imposed & A dozen equally-weighted buttons, no primary action \\
  C5 & Affordance           & Interactive elements look interactive & Clickable card with no hover/cursor cue \\
  C6 & Fitts's Law          & Primary actions large and near their context & Tiny submit button placed far from its form \\
  D1 & Spacing Consistency  & Even gaps between same-level elements & Card grid with jittery, unequal spacing \\
  D2 & Visual Balance       & Visual weight distributed, not lopsided & Content crammed left; right half dead space \\
  D3 & Content-Container Fit & Content fits without overflow / clip / under-fill & Heading text spilling out of its button \\
\end{tblr}
\caption{The 19-principle violation taxonomy: each principle, its definition, and an example of a page that violates it. Families: A accessibility (WCAG 2.2), B dark patterns, C cognitive and perceptual laws, D visual composition. A1--C6 are detectable from the HTML source; D1--D3 require the rendered screenshot.}
\label{tab:defs}
\end{table*}

\section{Data Pipeline Detail}
\label{app:pipeline}

This appendix expands the preparation pipeline summarized in Section~\ref{sec:data}.

\paragraph{Pass 1: structural eligibility.} We parse each page with BeautifulSoup and compute a binary eligibility vector $\mathbf{e}\in\{0,1\}^{19}$ (Table~\ref{tab:eligibility}). Conditions fall in three classes: \emph{style-dependent} (A1, A3, B3, C5, C6 need color/size/spacing readable from Tailwind classes or inline styles), \emph{element-count} (B2, C1 need $\geq$2 same-function elements; C2, C3, C4 need $\geq$3 sections or $\geq$7 items), and \emph{structural-presence} (the rest need a specific element type present). Because we regenerate every page as a self-contained Tailwind document (Pass~4), style-dependent conditions are satisfied by construction; the binding constraints in practice are the element-count and structural-presence conditions, which depend on the page's content.

\paragraph{Pass 1b: language and length.} Restrict to English (WebCode2M is $\sim$47\% English); drop the top 5\% by length (malformed outliers) and keep the top 40\% of the remainder (longer pages are structurally richer).

\paragraph{Pass 2: categorization.} Qwen3-8B (vLLM, no thinking) labels each page by domain (20 categories) and page type (20 categories) from its first 8K characters.

\paragraph{Pass 3: stratified subsampling.} WebCode2M skews to Article/Post (58.8\%) and Homepage (15.0\%). We sample equally across page types (capping common, taking all rare) then round-robin across domains, yielding a balanced 10K working set (scalable to 100K).

\paragraph{Pass 4: Tailwind regeneration.} A three-round Qwen3-8B process: (1) a domain-specific structural plan; (2) self-contained HTML generation conditioned on the plan plus sampled seeds (a color palette of 100, layout pattern of 95, and ``vibe'' of 90) for diversity; (3) domain-specific enhancement adding 2--4 sections. Output is 2.5K--40K characters (median 12.5K); pages over 25K tokens are dropped.

\paragraph{Pass 5: Rendering.} Headless Chromium (Playwright) renders each page at $1200\times1500$, capturing 1--4 screenshots over the scroll height, downscaled to $448\times544$ ($\sim$2.6 images/page on average).

\begin{table*}[t]
\centering
\small
\begin{tblr}{
  width = \textwidth,
  colspec = {l l X[l] c},
  hline{1} = {1.5pt, solid},
  hline{Z} = {1.5pt, solid},
  hline{2} = {0.8pt, solid},
  hline{7} = {0.4pt, solid},
  hline{12} = {0.4pt, solid},
  hline{18} = {0.4pt, solid},
  row{1} = {font=\bfseries},
  column{1} = {bg=gray!10},
}
  ID & Principle & Eligibility condition & Type \\
  A1 & Non-text Contrast & Tailwind color/border classes on $\geq$1 UI component, OR inline \code{style} with color/border & Style \\
  A2 & Semantic Structure & $\geq$1 semantic element (\code{<button>}, \code{<nav>}, \code{<a>}, \code{<h1>}--\code{<h6>}, \code{<label>}) present & Presence \\
  A3 & Target Size & $\geq$1 interactive element with Tailwind size/padding classes or inline dimensions & Style \\
  A4 & Page Language & \code{<html>} tag present & Presence \\
  A5 & Error Identification & $\geq$1 \code{<form>} containing $\geq$1 \code{<input>} & Presence \\
  B1 & Confirmshaming & $\geq$1 \code{<button>} or \code{<a>} element & Presence \\
  B2 & Misdirection & $\geq$2 action elements (buttons, submits, or CTA-classed links) & Count \\
  B3 & Hidden Costs & Tailwind text-size/color classes available, OR inline styles present & Style \\
  B4 & Trick Questions & $\geq$1 \code{<form>} with checkboxes/radios, OR $\geq$1 \code{<form>} where they can be added & Presence \\
  B5 & Forced Continuity & $\geq$1 \code{<button>} or \code{<a>} element & Presence \\
  C1 & Similarity & $\geq$2 same-function elements (buttons, nav links, cards) & Count \\
  C2 & Von Restorff & $\geq$3 sections with heterogeneous emphasis classes & Count \\
  C3 & Miller's Law & $\geq$7 items in any list (\code{<li>}), nav, form fields, or grid & Count \\
  C4 & Hick's Law & $\geq$7 equal-level choices (nav \code{<a>}, \code{<option>}, sibling cards) & Count \\
  C5 & Affordance & $\geq$1 interactive element with hover/shadow/cursor Tailwind classes & Style \\
  C6 & Fitts's Law & $\geq$1 primary action with Tailwind positioning/size classes & Style \\
  D1 & Spacing Consistency & $\geq$1 \code{gap-}, \code{space-y-}, or \code{mb-} spacing utility on repeated siblings & Style \\
  D2 & Visual Balance & Flex/grid layout or fractional-width (\code{w-1/}) columns present & Presence \\
  D3 & Content-Container Fit & $\geq$1 \code{<div>} container (text/element can be made to overflow or under-fill) & Presence \\
\end{tblr}
\caption{Pass 1 eligibility rules. Type indicates whether the condition depends on visible CSS (Style), element counts (Count), or mere element presence (Presence).}
\label{tab:eligibility}
\end{table*}

\section{Generation Categories}
\label{app:gen-categories}

Table~\ref{tab:categories} lists the 20 domain and 20 page-type categories used to classify seed pages (Section~\ref{sec:source-prep}). Stratified sampling across both axes covers categories that are rare in the source corpus but important for violation coverage, such as checkout and login pages.

\begin{table}[h]
\centering
\small
\begin{tblr}{
  colspec = {r l r l},
  hline{1} = {1.5pt, solid},
  hline{Z} = {1.5pt, solid},
  hline{2} = {0.8pt, solid},
  row{1} = {font=\bfseries},
}
  \# & Domain & \# & Page type \\
  1  & E-commerce / Retail   & 1  & Homepage / Landing \\
  2  & Social Media          & 2  & Article / Post \\
  3  & Gaming                & 3  & Product Page \\
  4  & Sports                & 4  & Login / Signup \\
  5  & News / Media          & 5  & Checkout / Cart \\
  6  & Education             & 6  & Pricing / Plans \\
  7  & Healthcare            & 7  & Settings / Account \\
  8  & Finance / Banking     & 8  & Contact / Support \\
  9  & Travel / Hospitality  & 9  & Dashboard / Admin \\
  10 & Food / Restaurant     & 10 & Search Results \\
  11 & Technology / SaaS     & 11 & Profile \\
  12 & Legal                 & 12 & Forum / Discussion \\
  13 & Government            & 13 & Documentation \\
  14 & Real Estate           & 14 & Directory / Listing \\
  15 & Entertainment         & 15 & Gallery / Portfolio \\
  16 & Fitness / Wellness    & 16 & Event / Calendar \\
  17 & Automotive            & 17 & Chat / Messaging \\
  18 & Non-profit            & 18 & FAQ / Help \\
  19 & Personal / Blog       & 19 & Download / Install \\
  20 & Other                 & 20 & Other \\
\end{tblr}
\caption{The 20 domain and 20 page-type categories used to classify seed pages and stratify sampling.}
\label{tab:categories}
\end{table}

\section{Prompts}
\label{app:gen-prompts}
\label{app:inject-prompts}

This appendix lists, verbatim, the generation prompts (Section~\ref{sec:source-prep}) and a representative injection/verification prompt from each of the four principle families (Section~\ref{sec:injection}). Fields in braces, e.g.\ \code{\{domain\}} and \code{\{html\_content\}}, are filled per page. The remaining principles use prompts of the same form and ship with the released recipe.


\subsection{Generation Prompts}
\label{app:gen-prompts-detail}
The three teacher rounds that regenerate each seed page.

\begin{promptlisting}{Round 1: Plan}
You are a senior UI/UX designer. You have been given the HTML of a real {page_type} page from a {domain} website.

Your job: analyze this page and describe exactly how to rebuild it as a professional, modern {domain} {page_type} page. DO NOT output any HTML. Only output a detailed plan.

STEP 1: EXTRACT THE SOURCE PAGE'S VISUAL IDENTITY
Look at the <style> block in the source HTML. Find all colors (hex codes, rgb values, color names). List them. These are the SOURCE COLORS -- your rebuilt page should use these as a BASE and modernize them.

Source colors you found: [list them]
Translated to Tailwind equivalents: [map each to closest Tailwind class]

Now BLEND the source colors with this style direction:
COLOR PALETTE: {color_seed}
VIBE: {vibe_seed}

Your final color choices should feel like the source page's identity modernized with the above direction. Not a copy, not a replacement -- a blend.

STEP 2: What does a real-world professional {domain} {page_type} page look like?
Think of the best examples you know of {domain} websites. What sections do they have? What layout patterns do they use? What interactive elements? What content structure? Be specific to {domain} -- a gaming forum looks NOTHING like a banking dashboard, a restaurant menu looks NOTHING like a SaaS pricing page.

STEP 3: LAYOUT STRUCTURE
Your page MUST use this layout pattern:
{layout_seed}

Describe how this layout pattern applies to a {domain} {page_type} page. What goes in each area?

STEP 4: What is in this source page?
List what the source page currently has: its sections, content, forms, buttons, navigation.

STEP 5: What is this page MISSING that a professional {domain} {page_type} page would have?
Be specific to this domain and page type. What do users expect? What actions do they want to take? What information do they need?

STEP 6: Write the full specification for the rebuilt page.
Describe every section from top to bottom. For each section specify:
- What it contains (exact text content, labels, numbers)
- How it's laid out (following the layout pattern from STEP 3)
- What interactive elements it has
- What colors it uses (from your blended palette in STEP 1)
- How it reflects the vibe: {vibe_seed}

The structure should match what {page_type} pages in {domain} actually look like in the real world, adapted to the specified layout pattern. NOT a generic template.

SOURCE PAGE:

{html_content}

YOUR PLAN:
\end{promptlisting}

\begin{promptlisting}{Round 2: Generate}
You are an expert frontend developer. You have been given a detailed design plan for a {page_type} page in the {domain} domain. Implement it EXACTLY as specified.

MANDATORY STYLE DIRECTIVES:
- COLOR PALETTE: {color_seed}
- LAYOUT: {layout_seed}
- VIBE: {vibe_seed}

You MUST use the colors specified above. Do NOT default to blue. If the palette says amber/stone, use amber/stone. If it says dark mode with neon, do that. The colors are non-negotiable.

You MUST follow the layout pattern specified. If it says "left sidebar with main content", do that -- not a centered single column. If it says "data table", use a table -- not cards.

CODE REQUIREMENTS:
- Start with <!DOCTYPE html> -- no markdown, no explanations
- Use <script src="https://cdn.tailwindcss.com"></script>
- Use semantic HTML5 elements (nav, main, section, article, aside, footer, header)
- All form inputs must have labels
- All images: <img src="https://placehold.co/WIDTHxHEIGHT" alt="description">
- All interactive elements must have hover/focus states using the palette colors
- Consistent spacing throughout

VISUAL REQUIREMENTS:
- Apply the VIBE throughout: {vibe_seed}
- The page must FEEL different from a generic template
- Use the exact color classes from the palette directive
- Layout must match the specified pattern -- NOT default centered cards

CONTENT REQUIREMENTS:
- Use realistic text from the plan (not lorem ipsum)
- All content should be {domain}-appropriate
- Match the content density specified in the plan

OUTPUT: Raw HTML only. Start with <!DOCTYPE html>, end with </html>. No markdown fences.

SOURCE PAGE FOR REFERENCE (use its content themes):

{source_html}

DESIGN PLAN TO IMPLEMENT:

{plan}
\end{promptlisting}

\begin{promptlisting}{Round 3: Enhance}
You are enhancing a {page_type} page for a {domain} website.

MANDATORY: Maintain these style directives throughout your enhancements:
- COLOR PALETTE: {color_seed}
- VIBE: {vibe_seed}

Do NOT change the existing color scheme. Your additions must use the SAME palette. If the page is dark with neon accents, your new sections are also dark with neon. If it's warm earth tones, stay warm.

YOUR TASK: Make this page significantly more complex and production-ready. Think about what a real {domain} {page_type} page has that this one is still missing. Add 2-4 more sections with domain-specific content and interactions.

RULES:
- Add content specific to {domain} -- not generic testimonials or FAQs
- Match existing color palette exactly -- use the same Tailwind classes already in the page
- Maintain the existing layout pattern -- don't restructure, just add depth
- Add at least 5 more interactive elements (appropriate to this page type)
- Add at least 100 more words of {domain}-specific content
- Do NOT remove anything -- only add
- Output COMPLETE page from <!DOCTYPE html> to </html>
- No markdown, no explanations -- raw HTML only

HERE IS THE CURRENT PAGE TO ENHANCE:

{generated_html}
\end{promptlisting}

\subsection{Representative Injection and Verification Prompts}
\label{app:inject-prompts-detail}
One injection/verification pair per family (A/B/C/D), verbatim.

\begin{promptlisting}{Injection: A1 Non-text Contrast}
You are creating a UI/UX examination page for students studying web accessibility. A student will need to identify the violation by examining the rendered page. Do not add HTML comments, placeholder text, dummy content, or any hints about what was changed.

ACCESSIBILITY VIOLATION: Non-text contrast failure (WCAG SC 1.4.11) means UI components such as borders, fills, focus indicators, and icons fail to maintain a minimum 3:1 contrast ratio against adjacent colors. Users with low vision cannot perceive component boundaries or interactive states when contrast is insufficient.

YOUR TASK: Find existing UI components on the page -- form inputs, buttons, cards, dividers, icons, or navigation elements -- and reduce their border/fill/icon contrast to below 3:1 against the adjacent background color. The components must still be structurally present and functional, just visually near-invisible due to insufficient contrast. Modify at least 3 distinct UI components.

CRITICAL: The violation must be VISIBLE at 420x530 screenshot resolution. Use EXTREME low contrast -- not "slightly faded" but "nearly invisible." The judge will compare screenshots and must clearly see that components have become ghost-like.

SPECIFIC TAILWIND CHANGES TO MAKE (use the most extreme options):

On white (bg-white) or very light backgrounds:
- Change border-gray-300 to border-white or border-gray-50 (effectively invisible)
- Change text-gray-700 (for icons) to text-gray-100 or text-white (icon disappears)
- Change bg-blue-600 (button fill) to bg-gray-50 or bg-white (button becomes invisible)
- Change divide-gray-200 to divide-white or divide-transparent
- Change ring-blue-500 to ring-white or ring-transparent

On bg-gray-50 or bg-slate-50 backgrounds:
- Change border-gray-300 to border-gray-50 (same as background = invisible)
- Change text-gray-600 (icons) to text-gray-100 (nearly invisible)
- Change bg-white cards to bg-gray-50 with border-gray-50 (card disappears into background)

On bg-gray-100 or bg-slate-100 backgrounds:
- Change border-gray-300 to border-slate-200 or border-gray-200
- Change text-gray-500 (icons) to text-gray-300

SPECIFIC COMPONENTS TO TARGET (find at least 3 on the page):
- Input fields: change border from border-gray-300 to border-gray-100 on bg-white
- Buttons: change border-gray-300 or bg-blue-600 to border-gray-100 or bg-blue-50 text-blue-100
- Card borders: change border-gray-200 to border-gray-50
- Icons inside buttons: change text-gray-500 to text-gray-200
- Dividers/separators: change border-gray-200 or divide-gray-200 to border-gray-50 or divide-gray-50
- Navigation links with underlines or indicators: change to near-background colors
- Toggle/switch components: change the track color to near-background
- Tab indicators: change active tab border-blue-600 to border-blue-100

DO:
- Modify at least 3 separate UI components
- Keep the components structurally intact (still in the DOM, still the right HTML elements)
- Use adjacent color pairs that technically render differently but fail the 3:1 contrast ratio
- Target borders, fills, and icons specifically (not text content -- that is covered by a different WCAG criterion)
- Make the changes subtle enough that a casual viewer might think elements are "missing" rather than low-contrast
- Preserve all text content at readable contrast (only reduce non-text component contrast)
- Keep hover and focus states equally low-contrast (change hover:border-gray-400 to hover:border-gray-200)

DON'T:
- Don't add HTML comments like <!-- low contrast --> or <!-- violation -->
- Don't remove components from the DOM entirely -- they must still be present, just hard to see
- Don't change text color to low contrast (this is specifically about non-text UI components: borders, fills, icons)
- Don't use opacity-0 or invisible -- components must render, just with insufficient contrast
- Don't change background colors of the page -- only change the component colors that sit on top of backgrounds
- Don't make only one component low-contrast -- must affect at least 3 elements
- Don't add any new elements to the page -- only modify existing ones
- Don't change the layout or spacing of any elements

OUTPUT FORMAT: Return the complete modified HTML followed by a paragraph (outside the code) explaining which components were modified, what their original and new color values are, and why each fails the 3:1 contrast ratio against its adjacent background.
\end{promptlisting}

\begin{promptlisting}{Verification: A1 Non-text Contrast}
You are verifying whether a non-text contrast accessibility violation (WCAG SC 1.4.11) was successfully injected into a web page. You will receive the original HTML, the modified HTML, a screenshot of the original page, and a screenshot of the modified page.

WHAT TO LOOK FOR: UI components (borders, fills, icons, focus indicators, dividers) that have been changed to colors with insufficient contrast (below 3:1) against their adjacent background. Components should appear to "disappear" or become very faint.

PASS CRITERIA (ALL must be met):
1. At least 3 distinct UI components have had their non-text visual properties (border, fill, icon color, divider) changed to lower-contrast values
2. The contrast between modified component colors and their backgrounds is clearly below 3:1 (e.g., border-gray-100 on bg-white, text-gray-200 icons on bg-white)
3. The components are still present in the DOM (not removed, not display:none)
4. The modification targets non-text elements specifically (borders, fills, icons), not body text content
5. In the modified screenshot, the affected components should appear significantly more faint or nearly invisible compared to the original screenshot

FAIL CRITERIA (any one means FAIL):
- Fewer than 3 components were modified
- The modified components still have clearly visible contrast (e.g., border-gray-400 on white is still roughly 3:1)
- Components were removed from the DOM rather than made low-contrast
- The modification primarily affected text content rather than UI component boundaries/fills/icons
- The contrast reduction is so minimal that components look identical in the screenshots
- opacity:0, invisible, or display:none was used instead of genuine low-contrast colors

Compare the screenshots carefully. In the original, you should see clearly defined borders, icons, and component boundaries. In the modified version, these same components should appear faded, ghostly, or nearly invisible against their backgrounds.

Answer YES if a genuine non-text contrast violation is present across at least 3 components. Answer NO if it fails any criteria. Provide a one-sentence justification referencing the specific components affected.
\end{promptlisting}

\begin{promptlisting}{Injection: B1 Confirmshaming}
You are creating a UI/UX examination page for students studying dark patterns. A student will need to identify the deceptive pattern by examining the rendered page. Do not add HTML comments, placeholder text, dummy content, or any hints about what was changed.

DARK PATTERN: Confirmshaming is a manipulative technique where the decline/opt-out option is worded with emotionally loaded, guilt-inducing language designed to shame the user into accepting the offer instead. The accept option is neutral and inviting, while the decline option implies the user is making a foolish, ignorant, or self-harmful choice by declining.

YOUR TASK: Find an existing call-to-action, modal, banner, popup, newsletter signup, discount offer, or similar conversion element on the page. Modify the DECLINE option so it uses guilt-tripping language. If no such element exists, add a contextually appropriate promotional banner or signup prompt that fits the page's existing theme and content.

EXAMPLE DECLINE PHRASES (choose one that fits the page context):
- "No thanks, I prefer to stay uninformed"
- "I don't care about saving money"
- "No, I enjoy paying full price"
- "I'll pass on improving my skills"
- "No thanks, I don't want to grow my business"
- "I'd rather miss out on exclusive deals"
- "No, I don't value my time"
- "I'll keep struggling on my own"
- "No thanks, I prefer the hard way"
- "I don't need expert guidance"

EXAMPLE ACCEPT PHRASES (keep these neutral/positive):
- "Yes, sign me up!"
- "Get my free guide"
- "Start saving today"
- "Join 50,000+ members"
- "Claim my discount"

STYLING REQUIREMENTS:
- The accept button: use classes like bg-blue-600 text-white px-6 py-3 rounded-lg font-semibold text-base or similar prominent styling that matches the page design
- The decline link: use classes like text-sm text-gray-500 mt-2 inline-block hover:text-gray-700 cursor-pointer -- it should look like a normal, understated text link
- The decline text must NOT be hidden or illegible -- it must be readable. The manipulation is in the WORDING, not the visibility
- The element must fit naturally into the page layout and content theme

DO:
- Match the guilt-trip language to the page's subject matter (fitness site = "No, I prefer being out of shape")
- Make the accept option sound appealing and neutral
- Use natural placement -- near existing CTAs, in a banner, or as a modal-style overlay
- Keep the surrounding content unchanged
- Use proper sentence structure in the decline text (capitalize first letter, end naturally)
- Make it look like a real marketing element that belongs on the page

DON'T:
- Don't add HTML comments like <!-- confirmshaming --> or <!-- violation -->
- Don't use obviously fake company names or placeholder text like "Company XYZ"
- Don't make the decline text invisible or illegible -- the dark pattern is in the LANGUAGE not visibility
- Don't use the word "confirmshaming" or "dark pattern" anywhere in the HTML
- Don't add a tooltip or aria-label that hints at the violation
- Don't change the overall page layout or remove existing content
- Don't make it look out of place -- it should blend with the existing design language

OUTPUT FORMAT: Return the complete modified HTML followed by a paragraph (outside the code) explaining what you added and where the student should look to find the confirmshaming language.
\end{promptlisting}

\begin{promptlisting}{Verification: B1 Confirmshaming}
You are verifying whether a confirmshaming dark pattern was successfully injected into a web page. You will receive the original HTML, the modified HTML, a screenshot of the original page, and a screenshot of the modified page.

WHAT TO LOOK FOR: Confirmshaming means the decline/opt-out option uses emotionally manipulative, guilt-inducing language that shames the user for declining. The accept option should be neutral/positive while the decline option implies the user is foolish, uninformed, or self-harming.

PASS CRITERIA (ALL must be met):
1. There exists a clear accept/decline choice pair on the modified page
2. The decline option uses guilt-inducing language (e.g., "No thanks, I prefer to stay uninformed" or "I don't care about saving money")
3. The guilt language is READABLE -- not hidden in invisible text or microscopic font
4. The accept option uses neutral or positive language (not also guilt-tripping)
5. The element fits contextually into the page (not random floating text)

FAIL CRITERIA (any one means FAIL):
- The decline option uses neutral language like "No thanks" or "Decline" or "Maybe later"
- The guilt-trip text is hidden (text-white on white, text-xs opacity-0, display:none)
- No accept/decline pair exists in the modified HTML
- The modification only added an HTML comment or invisible element
- The added text is clearly placeholder/dummy content (lorem ipsum, "Company XYZ offer")
- Both options use the same tone (both neutral or both aggressive)

Examine both the HTML source and the screenshots. Confirm the text is actually visible and readable in the screenshot.

Answer YES if a genuine confirmshaming dark pattern is present and visible to a user. Answer NO if it fails any criteria above. Provide a one-sentence justification.
\end{promptlisting}

\begin{promptlisting}{Injection: C5 Affordance}
You are creating a UI/UX examination page for students studying cognitive design laws. A student will need to identify the cognitive violation by examining the rendered page. Do not add HTML comments, placeholder text, dummy content, or any hints. Do not add section headers, dividers, or organizational elements that could chunk content.

VIOLATION: Affordance. Interactive elements (buttons, links, clickable cards) lack perceptual cues that signal clickability. They must look like plain, non-interactive text. A user should not be able to tell by visual inspection alone that these elements are clickable. The elements must still BE functional (still <button> or <a> in the DOM) but LOOK like static text.

WHY PREVIOUS ATTEMPTS FAILED: The LLM removed the background color but left hover:bg-blue-100, cursor-pointer, rounded-lg, and a visible border -- all of which still signal clickability. The LLM also sometimes only stripped 1-2 classes (e.g., removed bg-blue-600 but kept shadow-md, border, and rounded-lg which still make it look like a button). ALL visual affordance cues must be removed simultaneously. Another failure: the LLM only targeted ONE button; at least 3-4 interactive elements must lose their affordance.

STEP-BY-STEP INSTRUCTIONS:
1. Identify 3-6 interactive elements on the page that currently have clear button/link styling. Best targets: primary CTAs, navigation links with hover effects, card click areas, form submit buttons, toolbar buttons, tab elements.
2. For EACH targeted element, STRIP all of the following classes if present:
   - ALL background classes: bg-blue-600, bg-blue-500, bg-indigo-600, bg-green-500, bg-red-500, bg-white, bg-gray-100, bg-gray-200 -- remove ANY bg-* class entirely
   - ALL hover classes: hover:bg-*, hover:text-*, hover:shadow-*, hover:scale-*, hover:border-*, hover:opacity-*, hover:ring-* -- remove ANY hover:* class
   - ALL shadow classes: shadow-sm, shadow, shadow-md, shadow-lg, shadow-xl -- remove all
   - ALL border classes: border, border-2, border-gray-*, border-blue-*, border-transparent -- remove ANY border-* class
   - ALL rounded classes: rounded, rounded-md, rounded-lg, rounded-xl, rounded-full -- remove ALL rounded-* classes
   - cursor-pointer -- REMOVE this class
   - ALL ring classes: ring-*, focus:ring-* -- remove these
   - ALL transition/transform classes: transition, transition-all, transform, duration-* -- remove these
3. REPLACE with these classes to make it look like plain text:
   - ADD: cursor-default (signals non-interactivity)
   - ADD: text-gray-700 (plain text color, not a link color like text-blue-600)
   - KEEP: any text-base, text-sm, text-lg sizing class (normal text has size)
   - KEEP: any font-normal, font-medium (but remove font-semibold/bold if it makes it look like a CTA -- replace with font-normal)
   - REMOVE: text-white (this only made sense with a colored background)
   - REMOVE: uppercase, tracking-wide (these signal 'button' typography)
4. For <a> link elements specifically: also remove text-blue-600, text-blue-500, text-indigo-600, underline, hover:underline -- these are link affordance cues. Replace with text-gray-700 and no underline.
5. Verify the elements are still <button> or <a> tags in the DOM (do NOT change the tag to <span> or <div>). They must still be technically interactive.
6. Do NOT add pointer-events-none or disabled attributes -- the violation is VISUAL only. The element should function when clicked but not LOOK clickable.

DO:
- Strip bg-*, hover:*, shadow-*, border-*, rounded-*, cursor-pointer from at least 3-4 elements
- Add cursor-default and text-gray-700 to make elements look like plain text
- Remove text-white, uppercase, tracking-wide from targeted elements
- Remove font-semibold/font-bold from targeted elements (replace with font-normal)
- Keep the element tags as <button> or <a> (still functional in DOM)
- Make the buttons completely indistinguishable from surrounding paragraph text
- Target at least 3-4 interactive elements across the page
- Keep the rest of the page unchanged

DON'T:
- Add HTML comments
- Add dummy text or labels
- Leave ANY bg-* class on targeted elements
- Leave ANY hover:* class on targeted elements
- Leave ANY shadow-* class on targeted elements
- Leave ANY border-* or rounded-* class on targeted elements
- Leave cursor-pointer on targeted elements
- Change <button> to <span> or <div> (element must remain interactive in DOM)
- Add pointer-events-none or disabled attribute
- Only target ONE element (must be at least 3-4)
- Leave text-blue-600 or underline on links (these signal clickability)
- Keep font-semibold/bold on targeted buttons (makes them look like CTAs)

OUTPUT FORMAT:
First, output the complete modified HTML (from <!DOCTYPE html> to </html>) with no markdown fencing. Then, after a blank line, write a single paragraph starting with 'EXPLANATION:' describing which interactive elements you targeted and confirming that all visual affordance cues (bg, hover, shadow, border, rounded, cursor-pointer) were removed from each.
\end{promptlisting}

\begin{promptlisting}{Verification: C5 Affordance}
You are verifying whether a UI violation injection was performed correctly. The violation type is C6_affordance: interactive elements must LACK visual cues that signal clickability -- they should look like plain text.

You are given:
- ORIGINAL HTML: the clean page before injection
- MODIFIED HTML: the page after injection attempt
- ORIGINAL SCREENSHOT: rendered clean page
- MODIFIED SCREENSHOT: rendered modified page

WHAT TO CHECK:
1. ELEMENT COUNT: Identify how many interactive elements (button, a styled as button) were targeted. There must be at least 3-4 elements that lost their affordance.
2. For EACH targeted element, verify ALL of the following are ABSENT:
   - Background classes: bg-blue-*, bg-green-*, bg-indigo-*, bg-red-*, bg-gray-100, bg-gray-200, bg-white (when used for button fill), or ANY bg-* class
   - Hover classes: ANY class starting with hover:*
   - Shadow classes: shadow-sm, shadow, shadow-md, shadow-lg, shadow-xl
   - Border classes: border, border-2, border-*, or ANY border class
   - Rounded classes: rounded, rounded-md, rounded-lg, rounded-xl, rounded-full
   - cursor-pointer
   - Ring/focus classes: ring-*, focus:ring-*
3. Verify the targeted elements HAVE:
   - cursor-default (or no cursor class, which defaults to auto)
   - text-gray-700 or similar neutral text color (NOT text-blue-*, text-white)
   - font-normal (NOT font-semibold or font-bold)
4. Verify the elements are STILL <button> or <a> tags (not changed to <span>/<div>).
5. Verify no pointer-events-none or disabled attribute was added.
6. SCREENSHOT CHECK: The targeted elements should be visually indistinguishable from surrounding non-interactive text. They should NOT look clickable.
7. No HTML comments or hints were added.

PASS CRITERIA (ALL must be true):
- At least 3 interactive elements have been stripped of visual affordance
- Zero bg-*, hover:*, shadow-*, border-*, rounded-* classes on targeted elements
- cursor-pointer is absent from targeted elements
- Elements still use <button> or <a> tags (functionally interactive)
- No pointer-events-none or disabled attribute
- The screenshot shows buttons that look like plain text
- No HTML comments or dummy text added
- Rest of page is largely unchanged

FAIL CRITERIA (ANY triggers failure):
- Fewer than 3 elements were targeted
- ANY targeted element still has bg-*, hover:*, shadow-*, border-*, or rounded-*
- cursor-pointer remains on a targeted element
- The element tag was changed from <button>/<a> to <span>/<div>
- pointer-events-none or disabled was added
- In the screenshot, the targeted elements still look like buttons (visible background, border, shadow, or rounded shape)
- HTML comments or hints were added
- Only 1-2 elements were targeted (need at least 3)

OUTPUT: Answer YES if the injection passes all criteria, or NO if it fails. Then provide a 2-3 sentence justification listing which elements were targeted, what classes remain on them, and whether they look like plain text in the screenshot.
\end{promptlisting}

\begin{promptlisting}{Injection: D3 Content-Container Fit}
You are creating a UI/UX examination page for students studying visual layout quality. A student will need to identify the layout violation by examining a SCREENSHOT of the rendered page at 420x530 pixels. The violation must be dramatic and obvious at this small resolution. Do not add HTML comments or any hints.

VIOLATION: D4_content_container_fit -- Content and its container must be dramatically mismatched. Either (A) a container is absurdly oversized for its content (e.g., a box taking up half the viewport height containing only one short line of text), or (B) content visibly overflows/clips outside its container (text or images breaking out of their bounds). The mismatch must be immediately obvious at 420x530 resolution.

STEP-BY-STEP INSTRUCTIONS:

Choose ONE of these two sub-strategies:

STRATEGY A -- Oversized container (excessive empty space):

1. Find a container that holds a small amount of content: a card with a title, a section header, a single paragraph, or a button group.

2. Add an explicit large height class to it: h-72 (288px), h-80 (320px), or h-96 (384px). At 530px viewport height, h-80 is more than half the visible page -- extremely obvious.

3. The container must have visible boundaries. If it does not already have a border or background color, add one: "border border-gray-200" or "bg-gray-50" so the student can SEE the container boundaries and the empty space inside.

4. The content inside must remain small -- do NOT add content to fill the space. The result should be: a tall visible box with one or two lines of text floating at the top (or centered), with massive empty space below.

5. Add "min-h-[320px]" as a backup if Tailwind h-80 might be overridden by other classes.

STRATEGY B -- Content overflow (text/images breaking container bounds):

1. Find a container with text content -- a paragraph, a heading, a table cell, or a card description.

2. Add "overflow-hidden h-8" (or h-10, h-12) to the container to make it too short. Then add "whitespace-nowrap" to the text so it cannot wrap -- it will extend horizontally and get clipped by the overflow-hidden boundary. The text should be visibly cut off mid-word.

3. OR: Add "max-h-12 overflow-hidden" to a container that has 5+ lines of text. The text will be brutally truncated -- only showing 1-2 lines with the rest clipped off.

4. OR: Add "w-32 overflow-visible whitespace-nowrap" to a container with a long sentence. The text will visibly extend OUTSIDE the container, overlapping adjacent elements.

5. For image overflow: add "w-full" to a container and place an image with "min-w-[600px]" inside -- the image will extend beyond the container's right edge.

6. The overflow or truncation must be DRAMATIC. Clipping one word is not enough -- at least 50% of the content should be hidden or overflowing.

IMPORTANT FOR BOTH STRATEGIES:
- The violation must be in the visible viewport (above the fold at 530px height)
- There must be a visible container boundary (border, background, or shadow) so the mismatch between content and container is obvious
- Do not add text that describes the violation (no "this box is too big" content)

DO:
- Use h-72, h-80, or h-96 for oversized containers (these are 288px, 320px, 384px -- massive at 530px viewport)
- Add visible container boundaries (border, bg-color) so the empty space is clearly INSIDE a box
- For overflow: use whitespace-nowrap + overflow-hidden + constrained height to create dramatic text clipping
- Make the mismatch impossible to miss -- the container should be absurdly oversized OR content should be obviously clipped
- Target containers that are in the main content area, not buried in the footer

DO NOT:
- Use h-32 or smaller -- at 530px viewport this might not look extreme enough
- Create overflow without a visible container boundary (overflow without a box outline is hard to spot)
- Add HTML comments or labels
- Change text content to hint at the violation
- Use subtle overflow (1-2 words clipped) -- the majority of content must be affected
- Apply the height to the entire page/body (which just adds scrolling, not a visible container mismatch)
- Create the violation in an element that is hidden or below the fold

OUTPUT FORMAT:
First, output the complete modified HTML (from <!DOCTYPE html> to </html>) with the violation injected.
Then on a new line, write "---EXPLANATION---" followed by a brief paragraph explaining which strategy you used (A or B), which element you modified, what classes you added, and why it creates an obvious content-container mismatch at 420x530.
\end{promptlisting}

\begin{promptlisting}{Verification: D3 Content-Container Fit}
You are verifying whether a UI violation was successfully injected into a web page by comparing two screenshots rendered at 420x530 pixels.

VIOLATION TO CHECK: D4_content_container_fit -- Content should be dramatically mismatched with its container. Either (A) a container is absurdly large with barely any content inside (huge empty box), or (B) content visibly overflows or is clipped by its container.

You are given two images:
- LEFT/FIRST image: The ORIGINAL clean page (content fits containers naturally)
- RIGHT/SECOND image: The MODIFIED page (should have obvious content-container mismatch)

WHAT TO LOOK FOR:

1. OVERSIZED CONTAINER (Strategy A): Look for a box, card, or section in the modified screenshot that is dramatically taller than needed for its content. Indicators:
   - A bordered/colored box that is at least 50% of the viewport height but contains only 1-2 lines of text
   - Massive whitespace INSIDE a visible container (not just page margins)
   - A container that previously fit its content snugly now has an absurd amount of empty space
   - The empty space should be clearly INSIDE a bounded element, not just general page whitespace

2. CONTENT OVERFLOW (Strategy B): Look for content that is visibly cut off or extending beyond its container. Indicators:
   - Text that ends abruptly mid-word (horizontal or vertical clipping)
   - An image or element that extends beyond a visible container boundary
   - Text overlapping with adjacent elements because it overflows
   - A card or section where the bottom clearly clips through content

3. Compare the original and modified screenshots. In the original, content should fit naturally inside its containers. In the modified, there should be an obvious mismatch -- either excessive empty space within a box, or content being cut off.

4. The violation must be immediately visible. At 420x530 resolution:
   - An oversized container of h-80 (320px) takes up ~60% of the viewport height -- it will dominate the page
   - Text clipped to only 1-2 visible lines from a previously 5-line paragraph is immediately noticeable

PASS CRITERIA (answer YES):
- EITHER: A container is visibly much too large for its content (at least 3-4x taller than needed), creating a large empty area with visible boundaries (border or background)
- OR: Content is clearly cut off, truncated, or overflowing its container boundaries
- The mismatch is immediately obvious when comparing to the original screenshot
- The violation is visible without scrolling

FAIL CRITERIA (answer NO):
- All content appears to fit its containers naturally in both screenshots
- The original and modified screenshots look the same
- There is whitespace but no visible container boundary around it (just page margins, which are normal)
- Content is slightly truncated but still looks like a design choice (e.g., "Read more..." truncation)
- Any container size difference is subtle (less than 2x the needed height)

Respond with exactly:
VERDICT: YES or NO
JUSTIFICATION: 1-2 sentences describing what you see. If YES, identify whether it is an oversized container or content overflow, point to the specific element, and describe the severity. If NO, explain why content-container fit appears normal.
\end{promptlisting}

\end{document}